# EVolutionary Independent DEtermiNistiC Explanation

Vincenzo Dentamaro, Paolo Giglio, Donato Impedovo, Giuseppe Pirlo

***Abstract***

The widespread use of artificial intelligence deep neural networks (DNNs) in fields such as medicine and engineering necessitates understanding their decision-making processes. Current explainability methods often produce inconsistent results and struggle to highlight essential signals influencing model inferences. This paper introduces the Evolutionary Independent Deterministic Explanation (EVIDENCE) theory, a novel approach offering a deterministic, model-independent method for extracting significant signals from black-box models.
EVIDENCE theory, grounded in robust mathematical formalization, is validated through empirical tests on diverse datasets, including COVID-19 audio diagnostics, Parkinson's disease voice recordings, and the George Tzanetakis music classification dataset (GTZAN). Practical applications of EVIDENCE include improving diagnostic accuracy in healthcare and enhancing audio signal analysis. For instance, in the COVID-19 use case, EVIDENCE-filtered spectrograms fed into a frozen Residual Network with 50 layers (ResNet50) improved precision by 32% for positive cases and increased the Area Under the Curve (AUC) by 16% compared to baseline models. For Parkinson's disease classification, EVIDENCE achieved near-perfect precision and sensitivity, with a macro average F1-Score of 0.997. In the GTZAN, EVIDENCE maintained a high AUC of 0.996, demonstrating its efficacy in filtering relevant features for accurate genre classification.
EVIDENCE outperformed other Explainable Artificial Intelligence (XAI) methods such as Local Interpretable Model-agnostic Explanations (LIME), SHapley Additive exPlanations (SHAP), and Gradient-weighted Class-Activation Mapping (GradCAM) in almost all metrics. These findings indicate that EVIDENCE not only improves classification accuracy but also provides a transparent and reproducible explanation mechanism, crucial for advancing the trustworthiness and applicability of AI systems in real-world settings.

## I. INTRODUCTION AND RELATED WORK

The need of explainability has risen quickly in the last decades due to the pervasive implementation of AI algorithms to address a variety of practical tasks, [1]. Indeed, the need to explain how an algorithm took a specific decision or what it has found as meaningful in the input, is of paramount importance in fields where the consequences of such a decision is critical, like in the healthcare field, [2]. In these situations, it is a requirement the cooperation between professionals, such as doctors, nurses, scientists etc., and machines. Even if an algorithm seems to be able to perform better than the human capabilities on checking the presence of tiny nonetheless crucial details of information, the issue of generalization of the algorithms applicability on brand new data is nearly an indelible aspect of AI algorithms thus far. Consequently, professionals must constantly check the performances of machines when predicting or classifying information, to avoid inaccurate decisions or, in the worst case, false negative clinical assertions. Nevertheless, their professional judgement shall not be declassified with respect to what algorithms suggest, indeed it should represent the leading line of action supported with additional information. In this regard, AI algorithms' role is that to provide this extra knowledge before taking a crucial decision. To accomplish on these tasks there is a fundamental prerequisite: the true chance of understanding why a certain result was obtained and what information in terms of features and details was responsible for that outcome. This process would allow professionals, like doctors etc., to verify the reliability of the additional information provided by machines in cases where there is or there is not accordance with their own conclusions.
This is the role of Explainable Artificial Intelligence (XAI) systems. XAI refers to artificial intelligence (AI) systems whose decision-making processes can be transparently understood by humans, [3]. Trust in such decisions is involved when important consequences are at stake. In this regard, at present day it is important to justify a choice rather than another, for example on medical therapies, apart from the objective accuracy of the model that provided it, [2]. The professionals, such as a physician, need to check the "why" in this exchange of information with the machine for what would otherwise become an unbalanced relationship of passive execution for them. This issue is of undelayable attention due to the rapid advancement of AI ability to act on uncountable fields, [4].
On top of that, for any application on research advancement where there is not an "a priori knowledge", researchers doubly suffer ignorance, as they live in a space of no or partial understanding of what they are researching, neither they have an explanation of the AI model outcomes on that topic. For researching fields such as searching for new materials, researchers must test outcomes every time, [5]. Even though AI guesses it right, lack of model decision process' transparency keeps the knowledge advance stuck in blindness of the "why".

Apart from the centrality of professionals in taking the final responsibility for any decision, trust on AI models can't be taken for granted at present day. Biases are around the corner and can come up unexpectedly. Humans just can't afford complete trust in AI algorithms. A state-of-art XAI system for image classification might be trained to attend to the edges and textures of an image when recognizing objects, and then use attention weights to identify regions of the input image that mostly contributed to the model decision. This can help to understand if any bias is acting "behind the scenes" making the model focalize on irrelevant or misleading features, [6]. Potentially, something similar can happen in any field of AI application. Moreover, XAI systems can represent a strong tool to spot precious information from input data that helps to enhance accuracies due to hidden correlations that are not intelligible to a human brain.

A broad array of explainable artificial intelligence (XAI) algorithms has evolved over recent decades, showcasing varied methodologies, differing accuracies in isolating crucial data, and diverse analytical approaches to assessing the relevance of input features—both holistically and at granular levels. These techniques provide insightful analyses but vary significantly in their operational frameworks and dependencies, thus requiring meticulous classification for a thorough understanding of their efficacy and limitations.

A significant criterion for classifying XAI algorithms relies on the specific AI model architectures. Independent XAI methodologies, such as SHAP, in theory, operate without dependence on particular model structures, seeking to establish feature-output correlations irrespective of the predictive system's internal mechanisms, even if the practical implementation make distinctions among Tree based models and deep learning models. In contrast, dependent XAI methods, including GradCAM, are inherently linked to specific model designs. They achieve this by exploiting the structural and operational attributes of the prediction models to trace their decision pathways. Although such reliance facilitates targeted insights, it restricts their applicability across different architectures.

Despite the quantity of research in XAI methodologies in such a limited time-frame, a persistent challenge is the non-determinism. As highlighted in prior research, algorithms like LIME can produce inconsistent outcomes across repeated executions, even when applied to the exactly identical datasets and models. Furthermore, various algorithms often provide conflicting interpretations of the same inputs, especially concerning which patterns or features influenced the model's decision-making and are susceptible of hyperparameters optimization [60].

Existing methods, while valuable, exhibit distinct trade-offs. Perturbation-based approaches like LIME rely on local approximations, often leading to variations across runs. Gradient-based methods such as GradCAM depend heavily on specific architectures, which limits their generalizability. SHAP, while grounded in robust mathematical principles, can struggle with computational efficiency in high-dimensional scenarios. These limitations highlight the gap that EVIDENCE aims to fill by ensuring deterministic outputs and offering broad applicability across architectures and data types.

This inconsistency underscores the need for more reliable and deterministic XAI approaches that meet two essential criteria:
(1) Model's explanations must be deterministic, ensuring consistent results across runs for a given model and input;
(2) outputs should include only the critical features necessary for the model's inference.

In order to address the previously mentioned challenges, this work presents the Evolutionary Independent Deterministic Explanation model (EVIDENCE) model-independent explainable artificial intelligence theory and algorithm. EVIDENCE strides to provide a robust and mathematically proved theory of convergency to extrapolate all and only the signals that are recognized as important by the model, deterministically. It competes with other mathematically grounded methods such as Shap, which uses a game theoretical approach for deciding the most important features. EVIDENCE is designed to work with time-varying signals (such as audio) but can be extended also to other different type of unstructured signals such as 2D/3D images or videos.

The paper is organized as follows: Section II sketches the state of the art. Section III describes the EVIDENCE method and its convergence proof. Section IV sketches experimental results and discussion. Conclusions are presented in Section V.

## II. STATE OF THE ART

This section provides an in-depth review of existing XAI models and algorithms, organized to improve clarity and readability.

XAI algorithms can be categorized based on their approach, functionality, and dependency on AI models. Table 1 summarizes the main architectures used for various machine learning problems, while Table 2 provides a categorization of algorithms based on their intrinsic dependencies from the AI models. A brief summary of functionalities of the various XAI algorithms is presented in Table 3.

**Table 1** Overview of Explainable AI Models and Algorithms

| Model/Algorithm | Description | References |
|---|---|---|
| **SHapley Additive exPlanations SHAP** | Based on Shapley values from game theory, it highlights each feature's impact on the final prediction. | [7] |
| **Local Interpretable Model-agnostic Explanations LIME** | Fits a simpler interpretable model to AI predictions within a local domain of the input. | [8][9] |
| **Decision Trees** | Converts input data into a tree-like structure, making predictions retraceable. | [10] |
| **Bayesian Networks** | Uses probabilistic graphical models and Bayesian inference for transparent correlations. | [11][12] |
| **Counterfactual Explanations** | Generates answers to "what-if" questions to show alternative model behaviours. | [13] |
| **Attention-based Models** | Uses attention mechanisms to highlight impacting features, particularly in image analysis. | [14-16] |

**Table 2** Categorization of XAI algorithms based on their dependency on AI models

| Classification Type | Description | Example Models | References |
|---|---|---|---|
| **Model-Specific** | Dependent on specific AI model architectures for explainability. | Gradient-weighted Class Activation Mapping - GradCAM, GradCAM++ | [28-32] |
| **Model-Agnostic** | Can be applied to any AI model regardless of its architecture. | SHAP, LIME | [7-9] |
| **Local XAI Algorithms** | Provide insights into specific task performance of the AI model. | Local Interpretable Model-Agnostic Explanation (LIME) | [37-39] |
| **Global XAI Algorithms** | Aim to explain the entire AI model's working process. | Integrated Gradients | [19] |

**Table 3** Categories of Explainable AI Functionality and Example Techniques

| Functionality Category | Description | Example Models | References |
|---|---|---|---|
| **Black Box AI Model Explainers** | Post-hoc methods explaining already trained AI models. | Gradients, Integrated Gradients, DeepLIFT | [18-22] |
| **White Box Model Creators** | Focus on revealing how AI models make predictions. | - | [44-46] |
| **Equity Promoters** | Address biases in AI model decision processes to ensure fair predictions. | Fairness Constraints, Fair Representation Learning, Pre-processing and Post-processing Techniques | [47-52] |
| **AI Model Prevision Sensitivity Analyzers** | Analyse AI model sensitivity to input data changes, assessing robustness and reliability. | Local Sensitivity Analysis, Global Sensitivity Analysis, Perturbation-Based Sensitivity Analysis | [53-57] |

In recent years, the literature on XAI has expanded significantly, offering a plethora of models and methods designed to enhance the transparency and interpretability of AI systems. This section provides a comprehensive overview of the existing literature, categorizing the works based on their structural or input considerations.

- **Shapley Additive Explanations (SHAP)**

SHAP models utilise Shapley values from game theory to attribute the impact of each feature on the final prediction. This method roots itself in strong mathematical foundations, making it particularly robust for tabular data, though certain extensions exist for other data types [7].

- **Local Interpretable Model-Agnostic Explanations (LIME)**

LIME provides interpretability by fitting a simpler, interpretable model to the AI model's predictions within a local domain of the input. This approach is versatile and can be applied to any classification model, making it a widely used tool in XAI [8][9].

- **Decision Trees**

Decision tree-based models convert the original input data into smaller subsets through a cascade process, resulting in a tree-like structure. This allows for straightforward retracing of the input data that most significantly influenced the prediction [10].

- **Bayesian Networks**

Bayesian networks employ probabilistic graphical models and Bayesian inference to generate outcomes. Their simple, transparent architecture enables clear understanding of the correlations between input features and the resulting predictions [11][12].

- **Counterfactual Explanations**

Counterfactual explanations provide answers to hypothetical "what-if" scenarios, demonstrating how changes in input features would alter the model's behaviour. This approach is particularly useful for understanding the boundaries and conditions under which models operate [13].

- **Attention-based Models**

Attention-based models rely on mechanisms that focus on the most significant features for a given outcome. These models are especially effective for image analysis, as they can highlight specific features or regions of an image that contributed to the final decision [14-16].

- **Gradients and Related Techniques**

Gradient-based methods create class-specific saliency maps, which provide visual gradients of the input image based on the weight that areas had on the specific output of the AI algorithm. Techniques like Integrated Gradients and DeepLIFT extend this approach by relating model outputs to input features and enhancing the accuracy of neuron-level contributions [18-20].

- **Guided BackPropagation and Deconvolution**

Guided BackPropagation, also known as guided saliency, is designed to work with convolutional neural networks (CNNs). This method replaces max-pooling layers with convolutional ones to enhance the interpretability of feature activations. Similarly, Deconvolutional Networks (DCNNs) retrieve information on specific CNNs, enabling the understanding of input patterns that caused activations on feature maps [21-24].

- **Rise and Concept Activation Vectors (Tcav)**

Rise generates saliency maps by randomly masking the input image and measuring the related output multiple times, defining a pixel-related saliency distribution. Tcav links saliency map outputs to higher-level concepts understandable by users, enhancing interpretability [25-26].

- **Class Activation Maps (CAMs) and GradCAM**

CAMs are designed for CNNs to relate significant parts of the input image used for classification. GradCAM generalises this approach, providing gradient-based maps of input images that highlight crucial areas, regardless of the specific architecture [27-30]. GradCAM++ extends this further for multi-labelled problems [31-32].

- **Layer-wise Relevance Propagation (LRP) and SmoothGrad**

LRP decomposes nonlinear classifiers, especially deep neural networks, using backpropagation to identify meaningful input features. SmoothGrad, often used for denoising, works synergistically with other gradient-based methods to enhance interpretability by removing spurious information [33-36].

- **Local Interpretable Model-Agnostic Explanation (LIME) and Deterministic Lime (Dlime)**

LIME is model-agnostic and uses a perturbation approach to create a new dataset of input variants, explaining the AI model's decisions based on similarities with the original data. Dlime, developed to address LIME's non-deterministic nature, generates input variants deterministically [37-41].

- **Shapley Values and White Box Models**

SHAP, inspired by game theory, computes weights for features involved in predictions, while White Box models focus on revealing the AI model's decision-making process [42-46].

- **Equity Promoters and Sensitivity Analysers**

Equity promoters identify and address biases in AI models to ensure fair predictions. Techniques include fairness constraints, fairness representation learning, and pre- and post-processing methods. Sensitivity analysers evaluate AI model robustness and reliability by examining the impact of input data changes on model performance [47-57].

- **Adversarial Attacks**

Adversarial attacks involve making significant changes to input data to test AI models' robustness and identify vulnerabilities. This stress-testingrg approach helps uncover potential fragilities in AI models' decision-making processes [58-59].

An at a glance classification is depicted in Figure 1.

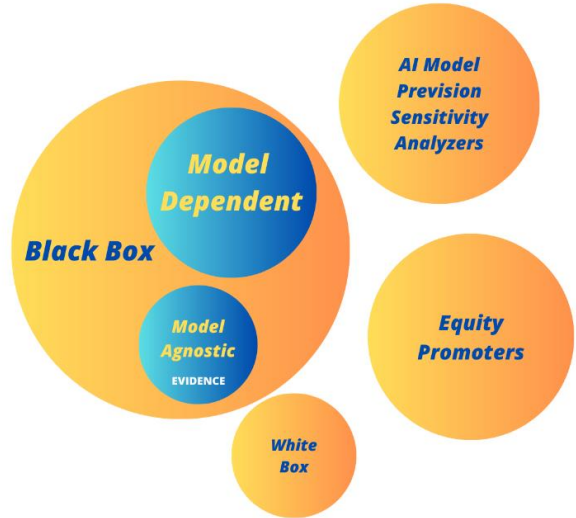

**Fig.1** Classification of XAI Algorithms

## II. METHODS

The overall work here proposed was developed into two moments of analysis: one focused on the introduction of a new explainable AI theory, named EVIDENCE, with its mathematical explanation; the other one focused on its application on a real case scenario study, finalized to compare its outcomes with other state-of-art explainable AI algorithms. While the latter analysis gives prompt results of the EVIDENCE application to allow for an easier evaluation of its efficacy, a mathematical description of its underlying logic is mandatory to give a strong foundation to its reliability.

### A. EVIDENCE: INTRODUCTION

Concerning the mathematical introduction of the EVIDENCE algorithm, the basic idea underneath its development is the generation of a population of signals from the input data, i.e., diversified inputs generated from the same original input signal. In this regard, every derived signal is obtained by keeping only a subset of the original information taken from the original input. The missing parts, those portions of information that are not taken from the original input, go to zero.
Consequently, an AI trained classification algorithm analyzes this diversified population and a score operator, such as the cross-entropy operator, is applied to the AI model classifications. Depending on the score operator outcomes, only a subset of the diversified population survives, which is the one exhibiting the best scores. That is equivalent to selecting only those diversified input data with a significative content of information, so that they allowed the AI classification model to perform consistently and to guess the right output. All this evaluation is possible due to the assumption of the presence of a ground truth for the original inputs. Another important assumption for this explainable algorithm to work is that of convergency of the subset of diversified population inputs to a single content of condensed information, which shall be the only information that resulted effective on allowing the AI algorithm to classify correctly.

### B. EVIDENCE: THE ALGORITHM

Here it is proposed the new EVIDENCE XAI deterministic algorithm. It was developed as a freestanding process designed to be independent from the typology of AI algorithm it should be applied on. EVIDENCE aims to keep the information that the deep neural network algorithm considers relevant for the classification purposes. The output of this algorithm can be considered a filtered version of the input, which in the next real case analysis will consist of a signal where only the most relevant features are still present. It is important to highlight that the algorithm is applied at the end of the learning process of the classification model, and it doesn't affect the training phase of the classification model, thus realizing that "independency" from the AI algorithm employed. In other words, EVIDENCE outcomes are independent from the specific AI model architecture, they are only related to the outcome performances of the classification process performed by the AI algorithm.

With more detail, the XAI algorithm pipeline starts with the AI model directly applied on a signal to be classified whose ground truth value is already known. For the mathematical description, a 2-dimensional Mel spectrogram of an audio track was considered as the input signal in the case study of this work; therefore, the EVIDENCE algorithm produces a population of different Mel Spectrograms that vary in terms of subcomponents, or chunks, of the original one. The deep learning classification model performs the classification task on these images. Then, the cross-entropy operator is applied between the model output and the ground truth value of the signal class. This operation allows finally to select those elements that present lower values of the cross entropy, i.e., a higher correlation, with respect to the expected classification result. The sub-population of higher correlated signals is then supposed to converge on the specific signal features that mostly contributed to the classification task. This process is equivalent to state that their linear combination is expected to converge to a finite sum, as it will be proven later.

*C. EVIDENCE: MATHEMATICAL DESCRIPTION*

It follows the mathematical description of the EVIDENCE algorithm along with a mathematical proof of its convergence on the desired filtered output of the most relevant features for the AI learning process.

**Mathematical description of EVIDENCE:** Let **M** be a $l \times d$ matrix of real numbers and let H be the Cross-Entropy operator, defined as in equation (1):

$$H: \mathbb{R}^2 \to \mathbb{R}, \quad H(p||q) \to -\sum_{x \in X} p(x) \cdot \log q(x) \tag{1}$$

Let $\psi$ be the frozen trained model (in this case a Deep Neural Network) used for classification purposes, whose internal decision rules satisfy the following requirements:
  a. Independency from the test input data,
  b. Independency from the cardinality of the input dataset.

$\psi$ receives the matrix **M** as an input and outputs a tuple of scores as in equation (2):

$$\psi(\mathbf{M}) = \text{scores} \leq 1, \text{with} \sum \text{scores} = 1 \tag{2}$$

To create filter masks to be applied on the matrix **M**, it is necessary to define their dimensionality, i.e., the freedom degrees of their variability, which is important for computational efficiency.

Thus, for a given natural number $m_0$, divisor of l, we can consider the generic vector **q** of dimensionality m defined as the tuple in (3):

$$q \in D'_{2,m} \text{ of elements } 1,0 \tag{3}$$

where $m = \frac{l}{m_0}$ and $D'_{2,m}$ is the set of dispositions with repetition of cardinality $2^m$ of the values 1 and 0. These dispositions represent the variability of the filter masks to be applied on the input matrix **M**.

Then, the generic filter vector **q** must be expanded to match the original row dimensionality of **M**: we can consider a set K of k functions defined as in equation (4) generating a vector **v** of length l by taking in input the generic vector **q** of length m:

$$k \in K \Leftrightarrow k: \mathbb{R}^m \to \mathbb{R}^l,$$

$$q_z \to \begin{cases} v_i = 1, \text{if } i \in [z \cdot m_0, (z+1) \cdot m_0] \wedge q_z \neq 0, 1 \leq z \leq m \\ v_i = 0 \text{ otherwise} \end{cases} \tag{4}$$

Finally, the following $l \times d$ filter matrix **F** is obtained as the augmentation of the $l \times 1$ row vector $\mathbf{v} = k(\mathbf{q})$:

$$\mathbf{F} \stackrel{\text{def}}{=} \mathbf{v} \otimes \mathbf{u} \tag{5}$$

with $\mathbf{u} = [1, \dots, 1]^T$ a $1 \times d$ column vector where all the elements are 1.

The Hadamard product matrix function can be defined as:

$$\mathbf{Q} \stackrel{\text{def}}{=} \mathbf{M} \circ \mathbf{F} \tag{6}$$

and it gives a filtered version of the original input matrix $\mathbf{M}$ by applying the filter matrix $\mathbf{F}$ on it.

By applying the $\psi$ model to the generic filtered input $\mathbf{Q}^c$, a tuple of scores is obtained, where every score represents the probability of the filtered input function $\mathbf{Q}^c$ to be classified as belonging to a certain class. Indeed, it is possible to apply the entropy operator H to the tuple of scores and the expected value of the outcome $E(\mathbf{M})$ in equation (7):

$$H(\psi(\mathbf{Q}^c) \| E(\mathbf{M})) = h_c \in \mathbb{R} \tag{7}$$

**Thesis:** considering the subset $\underline{Q}$ of cardinality n, with $n \leq 2^m$, of matrix functions defined in (6):
$\underline{\mathbf{Q}^c} \ni' h_c \leq W$, with $W \in \mathbb{R}$ defined as an arbitrary threshold value, it exists the function $\chi$ defined as in equation (7):

$$\chi = \lim_{n \to \infty} \frac{1}{n} \sum_{c=1}^{n} h'_c \cdot \underline{\mathbf{Q}^c} \tag{8}$$

with $h'_c = \frac{1}{h_c + 1}$.

**Proof:** Such an assessment is proposed as a consequence of the existence of the limit in (8). In this regard it should be observed that:

a. $\lim_{n \to \infty} \frac{1}{n} \sum_{c=1}^{n} h'_c \cdot \underline{\mathbf{Q}^c} \leq \lim_{n \to \infty} \frac{1}{n} \sum_{c=1}^{n} h'_c \cdot \max(\underline{\mathbf{Q}^c})$

and, by construction:

$$\lim_{n \to \infty} \frac{1}{n} \sum_{c=1}^{n} h'_c \cdot \max(\underline{\mathbf{Q}^c}) \leq \lim_{n \to \infty} \frac{1}{n} \sum_{c=1}^{n} h'_c \cdot \max(M) = \lim_{n \to \infty} \frac{1}{n} \cdot n \cdot h'_c \cdot \max(M) = h'_c \cdot \max(M)$$

thus:
$\lim_{n \to \infty} \frac{1}{n} \sum_{c=1}^{n} h'_c \cdot \underline{\mathbf{Q}^c} \leq h'_c \cdot \max(M)$.

Additionally,
$\lim_{n \to \infty} \frac{1}{n} \sum_{c=1}^{n} h'_c \cdot \underline{\mathbf{Q}^c} \geq \lim_{n \to \infty} \frac{1}{n} \sum_{c=1}^{n} h'_c \cdot \min(\underline{\mathbf{Q}^c})$

and, by construction:
$\lim_{n \to \infty} \frac{1}{n} \sum_{c=1}^{n} h'_c \cdot \min(\underline{\mathbf{Q}^c}) \geq \lim_{n \to \infty} \frac{1}{n} \sum_{c=1}^{n} h'_c \cdot \min(M) = \lim_{n \to \infty} \frac{1}{n} \cdot n \cdot h'_c \cdot \min(M) = h'_c \cdot \min(M)$

thus:
$\lim_{n \to \infty} \frac{1}{n} \sum_{c=1}^{n} h'_c \cdot \underline{\mathbf{Q}^c} \geq h'_c \cdot \min(M)$.

Therefore, the limit in (8) does not diverge:

$$h'_c \cdot \min(M) \leq \lim_{n \to \infty} \frac{1}{n} \sum_{c=1}^{n} h'_c \cdot \underline{\mathbf{Q}^c} \leq h'_c \cdot \max(M). \tag{9}$$

b. $\forall i, j: \underline{Q}^c_{i,j} \in \{M_{i,j}, 0\}$, by definition, i.e., $\underline{Q}^c_{i,j}$ can be 1 of those 2 values. Moreover, for an arbitrary cardinality n of the set

$$\underline{Q}: \sum_{c=1}^{n} \underline{Q}^c_{i,j} = \underline{N_n} \cdot M_{i,j} + \underline{R_n} \cdot 0 \tag{10}$$

with $\underline{N_n}, \underline{R_n} \in \mathbb{N} \ni' \underline{N_n} + \underline{R_n} = n$. $\underline{N_n}, \underline{R_n}$ represent the number of times $\underline{Q}^c_{i,j}$ is equal respectively to $M_{i,j}$ or 0. It is worth noting the following:

a. By hypothesis, the model $\psi$ is not dependent on the cardinality of $D'_{2,m}$.

  b. The model $\psi$ is not dependent on the cardinality n of the set $\underline{Q}$. On the contrary, the set $\underline{Q}$ is completely determined by $\psi$ with the condition $h_c \leq W$ applied on the set $D'_{2,m}$.

As the set $\underline{Q}$ is uniquely determined by $\psi$, so it is the ratio $A_{i,j} = \frac{N_n}{R_n}$. Therefore, $A_{i,j}$ is the result of the mere selection of matrix functions operated by $\psi$. As $\psi$ is independent from the cardinality n of $\underline{Q}$, there is no implicit dependency of $A_{i,j}$ from n.
Thus, the following stands:

$$\lim_{n \to \infty} \frac{1}{n} \sum_{c=1}^{n} h'_c \cdot \underline{Q^c}_{i,j} = \lim_{n \to \infty} \frac{1}{n} \cdot h'_c \cdot \left( \underline{N_n} \cdot M_{i,j} + \underline{R_n} \cdot 0 \right) = \lim_{n \to \infty} \frac{1}{n} \cdot h'_c \cdot \underline{N_n} \cdot M_{i,j}. \tag{11}$$

With:
$$\underline{N_n} = n - \underline{R_n} = \frac{A_{i,j}}{A_{i,j}+1} \cdot n = A_{i,j}' \cdot n,$$

We finally have $\forall\ i \in [1, l],\ \forall \in\ j\ [1, d]$:

$$\lim_{n \to \infty} \frac{1}{n} \sum_{c=1}^{n} h'_c \cdot \underline{Q^c}_{i,j} = \lim_{n \to \infty} \frac{1}{n} \cdot h'_c \cdot \underline{N_n} \cdot M_{i,j} = \lim_{n \to \infty} \frac{1}{n} \cdot h'_c \cdot A_{i,j}' \cdot n \cdot M_{i,j} = h'_c \cdot A_{i,j}' \cdot M_{i,j} \tag{12}$$

or, briefly:

$$\lim_{n \to \infty} \frac{1}{n} \sum_{c=1}^{n} h'_c \cdot \underline{Q^c}_{i,j} = h'_c \cdot A_{i,j}' \cdot M_{i,j} \tag{13}$$

where $A_{i,j}'$ represents a fraction of the overall set $\underline{Q}$ of functions $\underline{Q^c}$ that satisfy $\underline{Q^c}_{i,j} \neq 0$ for a given threshold W and a given model $\psi$, while $M_{i,j}$ is the input matrix value at point i,j.

Due to the theorem of the uniqueness of the limit, the limit in (8) exists and is therefore a unique value $\forall\ i \in [1, l],\ \forall \in\ j\ [1, d]$.
Equation (13) summarizes the overall meaning of EVIDENCE algorithm: the convergency of the set of functions $\underline{Q}$, i.e., the diversified inputs generated from the same original input signal **M**, results in a Hadamard product between the weights matrix **A′** and the same input **M**. This product is weighted by the scalar $h'_c$ which gives more importance to the inputs features resulting in a lower cross entropy. The matrix of likelihood weights **A′** represents the filtering activity of EVIDENCE, thus fulfilling its explainable task on the input matrix by modulating its values.

**Observations:** it is important to underline that the condition of $n \to \infty$ is an ideal case strictly dependent to the cardinality n of the set $\underline{Q}$ and cardinality $2^m$ of the set $D'_{2,m}$, with $m = \frac{1}{m_0}$. The feasibility of the previous theoretical assumptions is thus strictly dependent on how much the real case sampling conditions, i.e., the matrix l dimension, match those of the theoretical assumptions. With properly defined threshold W and unitary sampling window $m_0$, the conditions shall be easily matched due to the exponential increase of the $D'_{2,m}$ cardinality with respect to l.
The $\chi$ function is the result of filtering the raw input data in M by extrapolating only the features that mainly contributed to the best score performance of the deep neural network $\psi$ with respect to a given threshold value W for the Cross Entropy operator H applied on the outcomes. From these assumptions a further corollary can be obtained:

**Corollary:** the amplitudes of the function $\chi$ are a measure of classification importance of the components of the signal in M concerning what the operator $\psi$ finds relevant during its classification process, all this with respect to the outcome performance measured by applying the Cross Entropy operator H.
Indeed, by considering (11), the coefficient $A_{i,j}'$ represents the fraction of the overall set of function $\underline{Q}$ of cardinality n that satisfies $\underline{Q^c}_{i,j} \neq 0$. As $n \to \infty$, by the law of large numbers, the coefficients $A_{i,j}'$ tend to represent the frequency distribution of the l-domain values, i.e., their expected relevance in terms of contribution to define the $\chi$ function output, with $\chi$ function representing the most indicative features assessed by $\psi$ from M for its classification task.

What makes EVIDENCE an interesting alternative with respect to the state-of-art AI explainable algorithms is a reproducibility of the outcomes: the results are completely determined by the input data, the chosen model, the classification task to be performed.

Once those elements are determined, the outcome is definitive and objectively represent the parts of the signal that are decisive for the correct exploitation of the classification task by the specific model employed.

If there were no useful information at all, the convergency of the outcome would inevitably stabilize on a nearly constant value tending to zero. This would mean that there was no recurrent information to create a non-zero convergency pattern at all for the model right guesses with respect to that contained in the failures. A pictorial representation of the algorithm is presented in Figure 2.

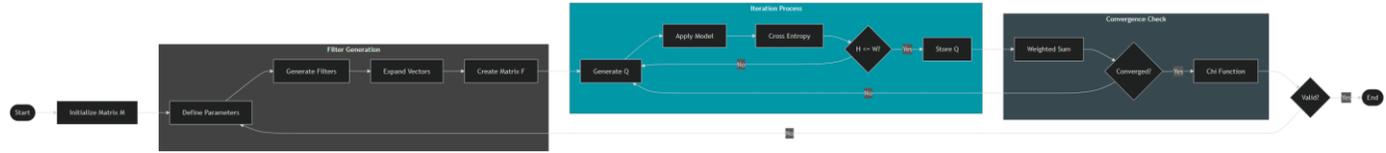

**Fig.2** Pictorial representation of the EVIDENCE algorithmic implementation

*D. CASE STUDIES*

EVIDENCE was tested on the following problems: (1) classification of Covid-19 PCR-Test positive users from healthy control ones without symptoms and with a clean medical history; (2) audio classification of people having Parkinson and Healthy Control subjects and (3) and the GTZAN 10 classes dataset.

For each problem, after the model was trained and tested, EVIDENCE was used on the frozen model, with the aim to filter the sound frequencies in the Mel Spectrograms that mostly contributed to a correct classification performance.
A similar process was performed by involving other two state-of-art explainable AI algorithms, specifically LIME, SHAP Deep Explainer (as suggested by the authors in [7] since the model is a CNN) and GRADCAM.
For each one of the explainability algorithms, the least relevant spectrograms features were removed based on their assessments, so that only the most significant ones were preserved. This was done for a reason: to let the already trained ResNet50 classify the input data once cleared from irrelevant information. This analysis shall estimate the explainability algorithms efficacy to select only valuable signal features for the classification task of the model, which is indirectly an explanation on what the trained ResNet50 considered important in the audio records with respect to their outcomes. The higher the accuracies of ResNet50, the higher the explainable algorithms' ability to filter only valuable information and explain with it what ResNet50 has learned for the task.
In general, regarding the following case studies, audio preprocessing involved resampling all recordings to 22050 Hz and normalizing amplitude levels. Mel spectrograms were generated using a window length of 2048 samples, a hop size of 344 samples, and 150 Mel filter banks. Each spectrogram was scaled to the decibel (dB) scale to enhance perceptual features. Silence trimming and zero-padding were applied where necessary. The process was implemented using the librosa Python library, ensuring consistent and reproducible results across datasets.

*Covid-19 Dataset*

The classification was performed by analyzing records of users' breaths and coughs, as it was demanded by a specific recording protocol. Indeed, the classification corresponds to the Task 1 of the Cambridge dataset [61,62].
The total cohort of the study is composed by 307 persons, 62 of them coming from positive Covid-19 PCR-Test and 245 for non-positive users. There were different audio tracks of breath and cough sounds for the same persons, thus in total there are 926 audio tracks: 282 audio tracks coming from Covid-19 Positive and 644 from healthy control subjects.
The algorithm employed for this task is a Deep Convolutional Neural Network model with a ResNet50 architecture. This model is trained with an inter-patient separation scheme, i.e. the dataset involved is built to keep the same proportion between positive and non-positive users' audio tracks both in the training and the test sets. This random procedure has been repeated 5 times in a 5-fold cross validation fashion and average results have been reported in Table 1.
If longer, audio tracks were cut-off at a time-length of 10 seconds. If shorter, the audio tracks were zero-padded to be 10 seconds long. A sampling rate of 22050 Hz was used. For each audio track, a Mel spectrogram was generated with the following parameters: 2048 bins of the FFT, 150 trainable filters, an overlapping window length of 140 milliseconds and a hop size of 344. The ResNet50 was trained from scratch and end-to-end on the generated spectrograms.
The test set consisted of audio tracks coming from 32 randomly chosen users, 16 Covid-19 PCR-Test positive users, representing the 25% of the total Covid-19 positive cohort and 16 healthy control subjects. Approximately 62 tracks from Covid-19 positive users and 40 tracks from healthy control subjects.

*Parkinson Dataset*

Voice recordings of the vowels /a/ and /i/ utilized in the study were collected as part of the research conducted by Hlavnicka et al. [63]. The study involved 83 Czech participants, comprising:

- 22 with Parkinson's disease (PD)
- 21 with Multiple System Atrophy (MSA)
- 18 with Progressive Supranuclear Palsy (PSP)
- 22 without any neurological disorders

The Unified Parkinson's Disease Rating Scale (UPDRS) was employed to measure disease severity, with trained neurologists assessing patients' motor skills. Specifically, UPDRS Part III was used to evaluate the severity of motor symptoms in PD patients. This part of the UPDRS ranges from 0 to 108 and assesses motor symptoms such as tremors, rigidity, bradykinesia, and postural stability. Higher scores indicate greater severity of motor symptoms. The average severity score for PD patients was 15.9, with a standard deviation of 7.9 [63].

Within the current sample, the participants included 22 PD patients (10 men and 12 women), 18 PSP patients (12 men and 6 women), 21 MSA patients (9 men and 12 women), and 22 healthy controls (11 men and 11 women). The mean ages were as follows: 64.4 years for PD patients (ranging from 48-82 years), 66.7 years for PSP patients (ranging from 54-84 years), 61.0 years for MSA patients (ranging from 45-71 years), and 63.6 years for the healthy control group (ranging from 41-79 years). The broader age range of the healthy control group reflects the age-related nature of idiopathic and atypical Parkinson's syndromes. Nevertheless, the mean ages are well-balanced, minimizing age-related confounding variables, thus ensuring the experiment's validity and the reliability of the results.

According to Hlavnicka et al. [63], all recordings were made in a low-noise environment using an Opus 55 condenser microphone positioned about 5 cm from the participants' lips. The recordings were digitized at a 16-bit resolution and a 48 kHz sampling rate. Each participant, guided by a trained specialist, was instructed to produce prolonged /a/ and /i/ vowels (in the international phonetic alphabet), maintaining a consistent modal voice. Each vowel was recorded at least twice, resulting in a total of 1011 recordings.

For the deep learning component, the audio signals were transformed into RGB-colored Mel-spectrogram images and then fed into neural network architectures like ResNet-50. The audio data was sampled at a rate of 22050 Hz. The Mel-spectrogram generation involved a Fast Fourier Transform (FFT) window length of 2048, a hop length of 344, and a Mel band count of 150. Additionally, the RMSProp algorithm was used as the gradient descent optimizer for the deep neural networks. RMSProp adjusts the weight update by computing the moving average of the squared gradient.

*GTZAN Dataset*

The GTZAN dataset comprises 1,000 audio song excerpts, each lasting 30 seconds, categorized into 10 different genres. These genres include but are not limited to rock, jazz, blues, and pop. [64] The dataset has been pivotal in the development and evaluation of machine learning models, including deep learning techniques, for audio classification tasks. Features such as mel-frequency cepstrum, tempo, and harmony have been extracted from the dataset at 3-second intervals. Also in this case for the deep learning aspect, audio signals are converted into RGB-colored Mel-spectrogram images and subsequently input into neural network architectures the one used is the as ResNet-50. The audio data is imported at a sampling rate of 22050 Hz. As before, the Mel-spectrogram images were used with a window length of 2048, a hop length of 344, and a Mel band count of 150. All audios were truncated at 3 seconds length and the RMSProp is used for the neural network optimization during gradient descent.

III. RESULTS & DISCUSSION

*COVID-19 Breath and Cough Sounds Classification*

The study aimed to evaluate the performance of the EVIDENCE explainable algorithm using a ResNet50 architecture trained through a 5-Fold cross-validation process with an inter-patient separation scheme. This setup was specialized to recognize COVID-19 patients based on the sound of their coughs and breathing. The performance metrics for a test set comprising 16 PCR-Test positive users and 16 PCR-Test negative users are detailed in Table 4 of Inferencing Results.

Ensuring the absence of biases in the input data and the learning process of the AI model was paramount. Four explainable algorithms EVIDENCE, LIME, GradCAM, and SHAP Deep Explainer, were employed to filter the relevant information from the ResNet50 model's perspective. These algorithms generated filtered versions of the test set, retaining only the non-trivial information according to each algorithm's assessment. The pre-trained ResNet50 model was then tested on these filtered test sets, and the outcomes are summarized in the table.

EVIDENCE demonstrated a substantial improvement over the baseline and other XAI methods across all metrics. Specifically, for COVID-19 negative cases, EVIDENCE increased precision from 0.86 (baseline) to 0.92, and sensitivity from 0.45 (baseline) to 0.94. For COVID-19 positive cases, EVIDENCE improved precision from 0.63 (baseline) to 0.95, while maintaining a high sensitivity of 0.90. The overall macro average F1-Score improved from 0.65 (baseline) to 0.94 with EVIDENCE, indicating a significant enhancement in recognizing COVID-19 patterns.

Conversely, LIME, GradCAM, and SHAP Deep Explainer displayed varying degrees of performance, generally lower than EVIDENCE. LIME exhibited a 13% drop in precision for negative cases compared to the baseline and a slight improvement in sensitivity for negative users. However, it resulted in an 11% decrease in sensitivity for positive users. The AUC for LIME was slightly lower than the baseline, indicating it did not retain all necessary patterns for correct classification.

GradCAM showed similar precision to the baseline but with lower sensitivity for negative cases. The AUC dropped by about 3%, suggesting the loss of important information during the filtering process. SHAP Deep Explainer performed similarly to GradCAM, but generally lower than both GradCAM and EVIDENCE in terms of AUC and F1-Score.

EVIDENCE's ability to enhance classification accuracy while filtering out non-significant information is noteworthy. Its deterministic nature ensures reproducibility and consistency in outcomes, making it a valuable tool for improving model transparency and reliability. The results highlight EVIDENCE's superior capability to identify and retain crucial information, leading to higher classification performance compared to other XAI methods.

*Parkinson Audio Classification*

In the context of Parkinson's disease classification, the EVIDENCE algorithm was tested using voice recordings of patients with Parkinson's, multiple system atrophy, and progressive supranuclear palsy, as well as healthy controls. The results in Table 4 indicated that EVIDENCE outperformed all other methods, achieving near-perfect precision and sensitivity for both healthy and Parkinson's cases. The macro average F1-Score for EVIDENCE was 0.997, significantly higher than the baseline and other XAI methods.
LIME and GradCAM exhibited notable drops in precision and sensitivity compared to the baseline. SHAP Deep Explainer had the lowest performance metrics, indicating its limited effectiveness in this application. The superior performance of EVIDENCE in this study underscores its robustness and reliability in filtering significant features while maintaining high classification accuracy.

*GTZAN 10 Classes Dataset*

In the GTZAN dataset, EVIDENCE was tested on audio excerpts from various music genres. The results showed that EVIDENCE maintained a high AUC of 0.996, significantly higher than other methods and the baseline. Although the precision and F1-Score were comparable to the baseline, the improvement in AUC demonstrates EVIDENCE's effectiveness in filtering and identifying relevant features for genre classification. LIME, GradCAM, and SHAP Deep Explainer showed lower performance metrics, with significant drops in precision and AUC compared to EVIDENCE. These results highlight the limitations of these methods in filtering meaningful information for accurate classification in this context.

Table 4 Inferencing results of the Mel-spectrograms generated by filtering the original spectrogram using the various XAI techniques. EVIDENCE outperformed all other techniques on Area Under the ROC Curve AUC metric

| Problem | XAI METHOD | Category | Precision | Sensitivity | F1-Score | Support | AUC |
|---|---|---|---|---|---|---|---|
| **Covid-19 breath and cough sounds classification** | BASELINE | Covid Negative | 0.86 | 0.45 | 0.57 | 16 | |
| | | Covid Positive | 0.63 | 0.90 | 0.73 | 16 | |
| | | Macro Average | 0.74 | 0.68 | 0.65 | 32 | 0.82 |
| | EVIDENCE | Covid Negative | 0.92 | 0.94 | 0.92 | 16 | |
| | | Covid Positive | 0.95 | 0.90 | 0.92 | 16 | |
| | | Macro Average | **0.94** | **0.92** | **0.94** | **32** | 0.99 |
| | LIME | Covid Negative | 0.73 | 0.49 | 0.56 | 16 | |
| | | Covid Positive | 0.62 | 0.79 | 0.68 | 16 | |
| | | Macro Average | 0.67 | 0.64 | 0.62 | 32 | 0.80 |
| | GRADCAM | Covid Negative | 0.88 | 0.49 | 0.58 | 16 | |
| | | Covid Positive | 0.66 | 0.89 | 0.74 | 16 | |
| | | Macro Average | 0.77 | 0.69 | 0.66 | 32 | 0.79 |
| | SHAP Deep Explainer | Covid Negative | 0.82 | 0.52 | 0.55 | 16 | |
| | | Covid Positive | 0.68 | 0.83 | 0.71 | 16 | |
| | | Macro Average | 0.75 | 0.72 | 0.66 | 32 | 0.74 |
| **Parkinson Audio Classification** | **BASELINE** | Healthy | 0.95 | 0.83 | 0.88 | 23 | |
| | | Parkinson | 0.85 | 0.96 | 0.90 | 23 | |
| | | Macro Average | 0.90 | 0.89 | 0.89 | 46 | 0.990 |
| | **EVIDENCE** | Healthy | **0.998** | **0.992** | **0.996** | 23 | |
| | | Parkinson | **0.999** | **0.995** | **0.997** | 23 | |
| | | Macro Average | **0.997** | **0.994** | **0.997** | 46 | 0.999 |
| | **LIME** | Healthy | 0.89 | 0.74 | 0.81 | 23 | |
| | | Parkinson | 0.78 | 0.91 | 0.84 | 23 | |
| | | Macro Average | 0.84 | 0.83 | 0.82 | 46 | 0.888 |
| | **GRADCAM** | Healthy | 0.999 | 0.13 | 0.23 | 23 | |
| | | Parkinson | 0.53 | 0.99 | 0.70 | 23 | |
| | | Macro Average | 0.77 | 0.57 | 0.46 | 46 | 0.809 |
| | **SHAP Deep Explainer** | Healthy | 0.75 | 0.39 | 0.51 | 23 | |
| | | Parkinson | 0.59 | 0.87 | 0.70 | 23 | |
| | | Macro Average | 0.67 | 0.63 | 0.61 | 46 | 0.742 |
| **GTZAN 10 classes** | **BASELINE** | Macro Average | 0.52 | 0.47 | 0.45 | 100 | 0.873 |
| | **EVIDENCE** | Macro Average | 0.51 | 0.61 | 0.51 | 100 | 0.996 |
| | **LIME** | Macro Average | 0.22 | 0.22 | 0.16 | 100 | 0.715 |
| | **GRADCAM** | Macro Average | 0.48 | 0.36 | 0.29 | 100 | 0.783 |
| | **SHAP Deep Explainer** | Macro Average | 0.49 | 0.50 | 0.47 | 100 | 0.816 |

Due to computing capacity, the application of the EVIDENCE theorem based on the convergence of non-trivial information into a filtered signal was stopped after 5000 iterations. For each test signal, the overall processing time amount was of about 48 seconds on average, with a multi-thread, GPU ready implementation. Once this operation was completed to return a filtered version of the test set, the already trained ResNet50 model was tested on it. The same procedure was adapted to perform with LIME, SHAP Deep Explainer and GradCAM on the same data, the respective outcomes of both the control test and the respective filtered versions of the test set are reported in Tab. 4.

Specifically, in all use cases, Lime was configured to extract a maximum of 100 features and 5000 iterations per image. These parameters were identified after hyperparameter tuning using Grid Search. GradCAM has just one parameter that is the target layer, in this case the target layer is named *conv5_block3_out* and was determined automatically by the GradCAM algorithm. Regarding SHAP Deep Explainer, it has been decided to discard the 50 percentile of Shap values, thus retaining only the remaining most important information. EVIDENCE was configured with 2 Hz chunks, 45 features and 5000 iterations for the Covid and Parkinson use cases, instead it was configured with 1 Hz chunk, 500 iterations and 200 features for the GTZAN problem.

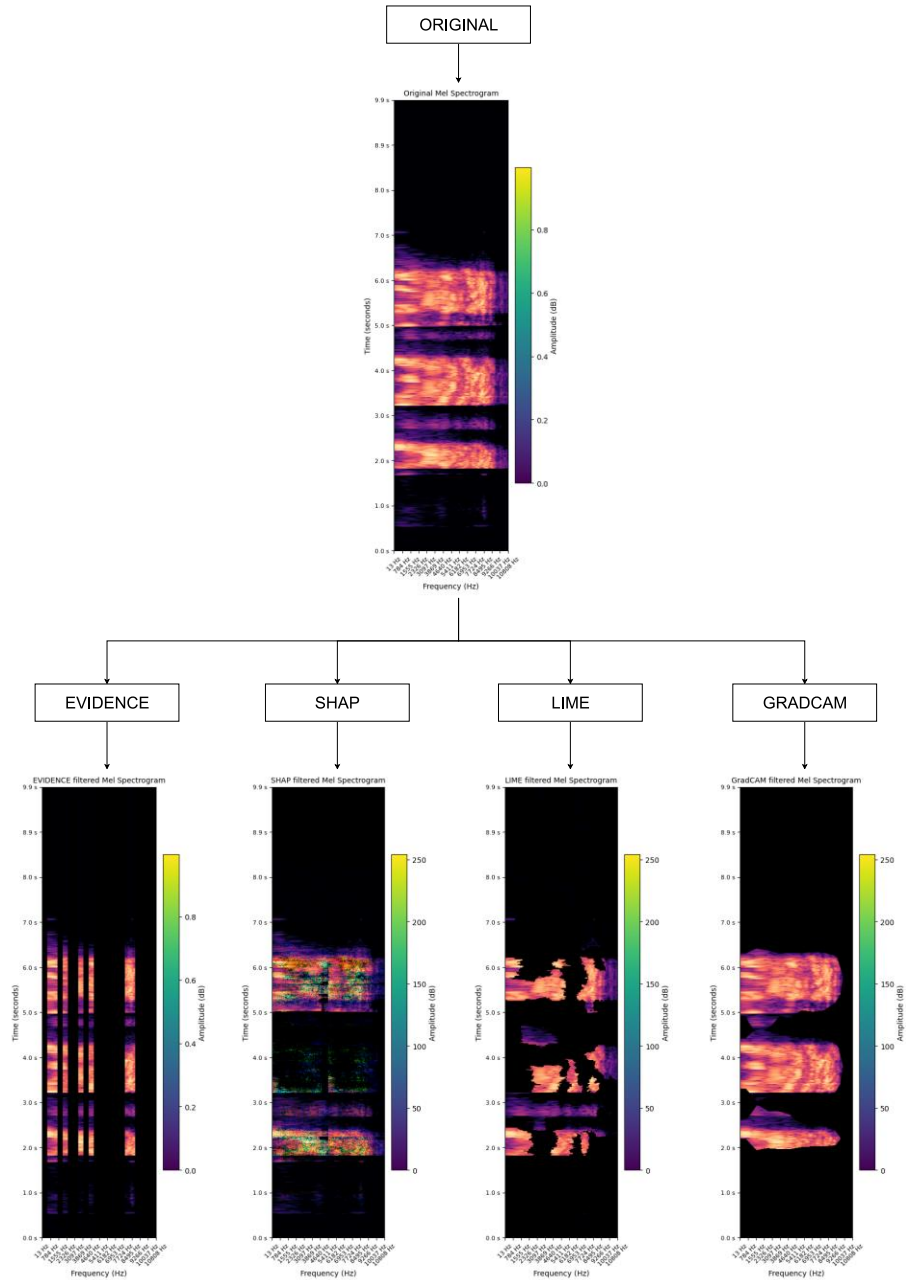

**Fig. 3** On top the original Mel Spectrogram of a covid positive breath sound. At the bottom from left to right, there is the EVIDENCE filtered Mel Spectrogram, the SHAP Deep Explainer filtered Mel Spectrogram, the LIME filtered Mel Spectrogram and the GRADCam Filtered Mel Spectrogram.

In all the use cases EVIDENCE algorithm was implemented downstream the whole pipeline to work on the spectrograms. The resulting filtering activity ended up with a cleaned version of the original spectrogram where all the trivial information was cut off. For each input item of the test set, the filtered outcome is the result of the convergency of a subpopulation of partial spectrograms of the same input. It is interesting to underline that, even though the partial spectrograms contain a percentage of trivial information, that shall gradually disappear in the convergence process due to its stochastic variability in the subpopulation. On the other hand, the non-trivial information shall gradually converge to a finite value, as shown with the mathematical formalization in the section II, due to its consistent presence in a not-negligible amount in the partial spectrograms population eligible to produce good classification accuracies.

LIME's underperformance derives from its local approximation approach, which attempts to explain individual predictions by creating locally interpretable models around each prediction by design. This inherently limits its ability to capture global patterns and relationships, especially when dealing with complex input data like spectrograms or high-dimensional data. In facts, LIME's local linear approximations can miss crucial non-linear relationships that determine model decisions. This limitation is particularly evident when features have strong interdependencies, as shown by its significant performance drops across all three audio classification tasks. GradCAM's primary limitation lies in its architecture-specific approach, focusing solely on the final convolutional layer's activations. While this works well for simple classification tasks, it struggles with complex data where discriminative features might be captured in earlier layers or through combinations of layer activations. Its reliance on class-specific gradients also means it might miss features that are jointly important for multiple classes, explaining its sharp sensitivity drops (e.g., 0.13 for healthy Parkinson's cases despite 0.999 precision). SHAP Deep Explainer's underperformance is rooted in its computational approximations necessary for handling deep networks. In order to maintain computational feasibility, it risks losing important feature interactions. Moreover, its additive feature attribution approach may not adequately capture complex non-linear relationships in the model's decision-making process, leading to the lowest overall performance metrics across different applications (macro average F1-scores consistently below baseline).

Regarding the Covid-19 use case, in Fig. 3 it is shown the original spectrogram (on the top) which is passed into the four different XAI methods and their resulting important parts of the spectrogram. The subpopulation of partial spectrograms was obtained with a stochastic filter parameter of 30%, i.e., each derived spectrogram contains only 30% of the original spectrogram information.

Differently from LIME, SHAP Deep Explainer and GradCAM, EVIDENCE performs a 1-dimensional filtering activity on frequencies. This is based on the hypothesis that no valuable information can be retrieved on the subjective timing of users' records of their coughs and breaths. The result was a spectrum with dark and bright bands for respectively significative and non-significative ranges of frequencies in the spectrograms. The most substantial outcome, anyway, is that the frozen ResNet50 model was able to achieve these increments of performances on the EVIDENCE filtered test set by relying on less than half of the original amount of information. Adding more information would gradually lead to the baseline case with some drop for the metrics. This, finally, is the proof that EVIDENCE was able to keep all and only the necessary information for the classification task, while filtering out the noisy one that would inevitably lead to performance drops.

Additionally, EVIDENCE can achieve these results by being totally independent from the AI classification model adopted, making it a generalizable and deterministic tool: its filtered outcomes of different AI models can be compared without any dependency from them. Finally, another interesting aspect is the following: EVIDENCE provides a definitive spectrogram outcome that is the result of the convergency of thousands of partial versions of the original input one. Along with convergency to non-trivial information, the corollary of the theorem suggests that some statistics can be performed on the set of partial input versions, thus allowing to identify the most important range of frequencies of the original signal that mainly contributed to the Covid-19 PCR-Test classification by ResNet50. Fig. 4 shows the convergency outcomes for both coughs and breaths spectrograms of a specific Covid-19 PCR-Test user.

Indeed, those frequencies above the threshold represent the signal components that mostly allowed the ResNet50 model to correctly guess Covid-19 PCR-Test positive users from negative ones. The resulting frequencies intervals are around 233 Hz to 674 Hz for breath sounds. While for coughs frequencies are in the intervals 10-894 Hz, 1335-1555 Hz, and around 1996-2447 Hz. Both plots show a high distribution towards low frequency spectrum. A reconstruction of the original audio signal filtered on these frequencies allowed to understand that those frequencies probably are correlated to language independent sounds with the potential to act as a universal gauge of the health condition of the patient. However, additional analysis will be required to verify this hypothesis.

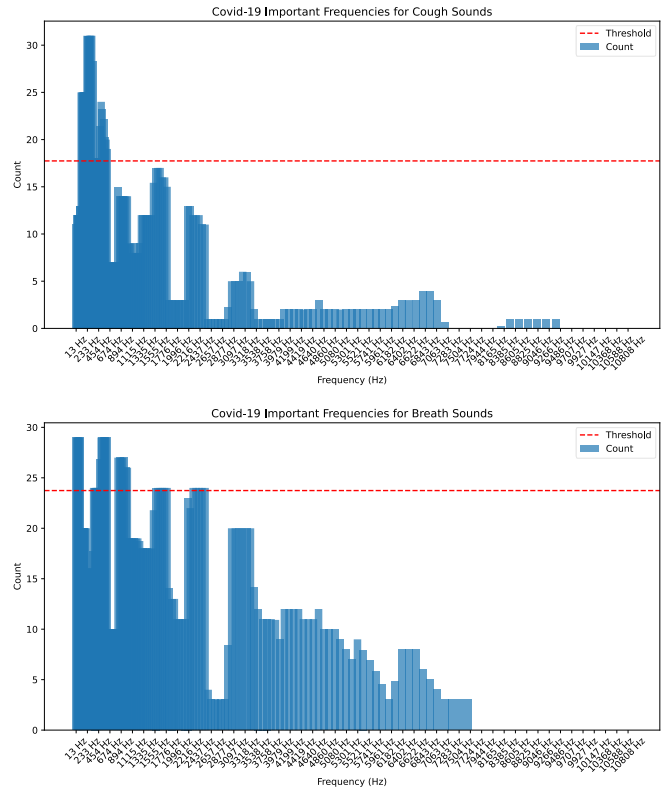

**Fig. 4** The most important range frequencies found in the Covid-19 use case on cough and breath spectrograms generated by EVIDENCE algorithm for coughs (top image) and breaths (bottom image). A trend in peaks of counts emerges which may be correlated to the spectrogram amplitudes of the respective frequencies. The dashed red line represents the average count of all the frequencies.

IV. CONCLUSIONS

EVIDENCE, a novel explainable AI algorithm, was devised with the notable advantage of being independent of the specific AI models employed for classification tasks. A robust mathematical formalisation was undertaken to substantiate its reliability. Beyond its theoretical underpinning, EVIDENCE was empirically tested across several datasets, including a ResNet50 model trained to identify COVID-19 affected individuals through the analysis of audio recordings of their coughs and breathing, voice recordings from patients with Parkinson's disease, and the GTZAN music genre dataset.

In the COVID-19 case study, EVIDENCE identified specific sounds that serve as indicators of COVID-19, demonstrating language-independent and potentially universal sound patterns. When juxtaposed with other explainable AI algorithms such as LIME, SHAP Deep Explainer, and GradCAM, EVIDENCE showed substantial improvements in classification outcomes. Specifically, it achieved a 32% enhancement in precision for COVID-19 PCR-test positive users and a 16% increase in the Area Under the Curve (AUC) for detecting COVID-19, relative to the baseline. Furthermore, EVIDENCE outperformed other state-of-the-art methods, evidencing a 19%, 25%, and 20% increase in AUC over LIME, SHAP Deep Explainer, and GradCAM, respectively. These enhancements underscore the ResNet50 model's augmented capability to detect COVID-19 PCR-Test positivity when utilising EVIDENCE-filtered data.

In the context of Parkinson's disease classification, EVIDENCE again outshone other methods. It achieved near-perfect precision and sensitivity for both healthy individuals and Parkinson's patients. The macro average F1-Score for EVIDENCE was 0.997, markedly higher than the baseline and other XAI methods. LIME and GradCAM exhibited notable reductions in precision and sensitivity relative to the baseline, while SHAP Deep Explainer recorded the lowest performance metrics. These findings underscore EVIDENCE's robustness and reliability in filtering significant features whilst maintaining high classification accuracy.

Regarding the GTZAN music genre classification task, EVIDENCE maintained an impressive AUC of 0.996, significantly surpassing other methods and the baseline. Although the precision and F1-Score were comparable to the baseline, the improvement in AUC underscores EVIDENCE's efficacy in filtering and identifying relevant features for genre classification. LIME, GradCAM, and SHAP Deep Explainer exhibited lower performance metrics, with significant reductions in precision and AUC compared to EVIDENCE. These outcomes highlight the limitations of these methods in filtering meaningful information for accurate classification in this context.

The exceptional performance of EVIDENCE across these diverse datasets suggests its potential to markedly enhance AI model classification accuracies. Its capacity to filter non-trivial information whilst cleansing input data is particularly noteworthy. Future analyses will aim to elucidate how EVIDENCE's filtering process contributes to these enhancements. This ongoing research aspires to further cement the precision of explainable AI algorithms in identifying critical information and preserving data integrity, thereby advancing the field of AI explainability and reliability.

Additionally, its deterministic and mathematical foundations of EVIDENCE make it fit for further extension in crucial domains. In financial implementations, it may increase the interpretability of trading strategies and fraud-detection systems and credit-scoring models, where legal requirements and transparency of the decision making processes is essential. As for autonomous driving, EVIDENCE may be able to scene enhancing and path planning decision making that helps in understanding vital aspects which forms the core of the AI decision making during critical safety parameters.

Nonetheless, a 48-second analysis with 5000 iterations, with the actual non optimized version of the algorithm, seems far from real time in its current form this for instance suggests an optimization opportunity. It is observed that the primary contributors to computational complexity are the matrix dimension of the input, the number of iterations, and the size of the sampling window. In active mode, efficient implementation would focus on optimizing parallel processing, lowering the number of iterations, and using accelerators such as GPUs. Future optimizations may involve adaptive iteration counting and stopping strategies which will enable real time performance without compromising on mathematical assurance of the methodology and its explanatory strength.

In future works, EVIDENCE will be extended to manage n-dimensional data, thus opening deterministic explainability doors to problems in many different domains from healthcare to security: it is possible to imagine a general EVIDENCE algorithm capable of deterministically explain AI patterns in image classification, 3D data, drug discovery and much more. Additionally, from a human computer interaction perspective, EVIDENCE could be integrated into visualization and machine learning de-biasing techniques.

ACKNOWLEDGMENT

This work was partially supported by the project FAIR - Future AI Research (PE00000013), spoke 6 - Symbiotic AI, under the NRRP MUR program funded by the NextGenerationEU.

REFERENCES

[1] Gupta, L.K., Koundal, D. and Mongia, S., 2023. Explainable Methods for Image-Based Deep Learning: A Review. *Archives of Computational Methods in Engineering*, pp.1-16
[2] Chaddad, A., Peng, J., Xu, J. and Bouridane, A., 2023. Survey of Explainable AI Techniques in Healthcare. *Sensors*, 23(2), p.634.
[3] Dwivedi, R., Dave, D., Naik, H., Singhal, S., Omer, R., Patel, P., Qian, B., Wen, Z., Shah, T., Morgan, G. and Ranjan, R., 2023. Explainable AI (XAI): Core ideas, techniques, and solutions. *ACM Computing Surveys*, 55(9), pp.1-33.
[4] Saeed, W. and Omlin, C., 2023. Explainable ai (xai): A systematic meta-survey of current challenges and future opportunities. *Knowledge-Based Systems*, p.110273.
[5] Zhong, X., Gallagher, B., Liu, S., Kailkhura, B., Hiszpanski, A. and Han, T.Y.J., 2022. Explainable machine learning in materials science. *npj Computational Materials*, 8(1), p.204.
[6] Hofeditz, L., Clausen, S., Rieß, A., Mirbabaie, M. and Stieglitz, S., 2022. Applying XAI to an AI-based system for candidate management to mitigate bias and discrimination in hiring. *Electronic Markets*, pp.1-27.
[7] Van den Broeck, G., Lykov, A., Schleich, M. and Suciu, D., 2022. On the tractability of SHAP explanations. *Journal of Artificial Intelligence Research*, 74, pp.851-886.
[8] Visani, G., Bagli, E., Chesani, F., Poluzzi, A. and Capuzzo, D., 2022. Statistical stability indices for LIME: Obtaining reliable explanations for machine learning models. *Journal of the Operational Research Society*, 73(1), pp.91-101.
[9] Hamilton, N., Webb, A., Wilder, M., Hendrickson, B., Blanck, M., Nelson, E., Roemer, W. and Havens, T.C., 2022, December. Enhancing Visualization and Explainability of Computer Vision Models with Local Interpretable Model-Agnostic Explanations (LIME). In *2022 IEEE Symposium Series on Computational Intelligence (SSCI)* (pp. 604-611). IEEE.
[10] Blanco-Justicia, A. and Domingo-Ferrer, J., 2019. Machine learning explainability through comprehensible decision trees. In *Machine Learning and Knowledge Extraction: Third IFIP TC 5, TC 12, WG 8.4, WG 8.9, WG 12.9 International Cross-Domain Conference, CD-MAKE 2019, Canterbury, UK, August 26–29, 2019, Proceedings 3* (pp. 15-26). Springer International Publishing.
[11] Derks, I.P. and De Waal, A., 2020. A taxonomy of explainable Bayesian networks. In *Artificial Intelligence Research: First Southern African Conference for AI Research, SACAIR 2020, Muldersdrift, South Africa, February 22-26, 2021, Proceedings 1* (pp. 220-235). Springer International Publishing.
[12] Valero-Leal, E., Larranaga, P. and Bielza, C., 2022, June. Extending MAP-independence for Bayesian network explainability. In *Proceedings of the First International Workshop on Heterodox Methods for Interpretable and Efficient AI*.
[13] Stepin, I., Alonso, J.M., Catala, A. and Pereira-Fariña, M., 2021. A survey of contrastive and counterfactual explanation generation methods for explainable artificial intelligence. *IEEE Access*, 9, pp.11974-12001.
[14] Shi, W., Tong, L., Zhuang, Y., Zhu, Y. and Wang, M.D., 2020, September. Exam: an explainable attention-based model for covid-19 automatic diagnosis. In *Proceedings of the 11th ACM international conference on bioinformatics, computational biology and health informatics* (pp. 1-6).
[15] Manica, M., Oskooei, A., Born, J., Subramanian, V., Sáez-Rodríguez, J. and Rodriguez Martinez, M., 2019. Toward explainable anticancer compound sensitivity prediction via multimodal attention-based convolutional encoders. *Molecular pharmaceutics*, 16(12), pp.4797-4806.
[16] Ahmed, U., Lin, J.C.W. and Srivastava, G., 2021, July. Fuzzy explainable attention-based deep active learning on mental-health data. In *2021 IEEE International Conference on Fuzzy Systems (FUZZ-IEEE)* (pp. 1-6). IEEE.
[17] Hariharan, S., Rejimol Robinson, R.R., Prasad, R.R., Thomas, C. and Balakrishnan, N., 2022. XAI for intrusion detection system: comparing explanations based on global and local scope. *Journal of Computer Virology and Hacking Techniques*, pp.1-23.
[18] Bodria, F., Giannotti, F., Guidotti, R., Naretto, F., Pedreschi, D. and Rinzivillo, S., 2021. Benchmarking and survey of explanation methods for black box models. *arXiv preprint arXiv:2102.13076*.

[19] Buhrmester, V., Münch, D. and Arens, M., 2021. Analysis of explainers of black box deep neural networks for computer vision: A survey. *Machine Learning and Knowledge Extraction*, *3*(4), pp.966-989.
[20] Machlev, R., Heistrene, L., Perl, M., Levy, K.Y., Belikov, J., Mannor, S. and Levron, Y., 2022. Explainable Artificial Intelligence (XAI) techniques for energy and power systems: Review, challenges and opportunities. *Energy and AI*, p.100169.
[21] Bhat, A., Assoa, A.S. and Raychowdhury, A., 2022, October. Gradient Backpropagation based Feature Attribution to Enable Explainable-AI on the Edge. In *2022 IFIP/IEEE 30th International Conference on Very Large Scale Integration (VLSI-SoC)* (pp. 1-6). IEEE.
[22] Van der Velden, B.H., Kuijf, H.J., Gilhuijs, K.G. and Viergever, M.A., 2022. Explainable artificial intelligence (XAI) in deep learning-based medical image analysis. *Medical Image Analysis*, p.102470.
[23] Das, A. and Rad, P., 2020. Opportunities and challenges in explainable artificial intelligence (xai): A survey. *arXiv preprint arXiv:2006.11371*.
[24] Jin, W., Li, X., Fatehi, M. and Hamarneh, G., 2023. Generating Post-Hoc Explanation from Deep Neural Networks for Multi-Modal Medical Image Analysis Tasks. *MethodsX*, p.102009.
[25] Li, X.H., Shi, Y., Li, H., Bai, W., Cao, C.C. and Chen, L., 2021, August. An experimental study of quantitative evaluations on saliency methods. In *Proceedings of the 27th ACM sigkdd conference on knowledge discovery & data mining* (pp. 3200-3208).
[26] Druc, S., Balu, A., Wooldridge, P., Krishnamurthy, A. and Sarkar, S., 2022. Concept Activation Vectors for Generating User-Defined 3D Shapes. In *Proceedings of the IEEE/CVF Conference on Computer Vision and Pattern Recognition* (pp. 2993-3000).
[27] Byun, S.Y. and Lee, W., 2022. Recipro-CAM: Gradient-free reciprocal class activation map. *arXiv preprint arXiv:2209.14074*.
[28] Joo, H.T. and Kim, K.J., 2019, August. Visualization of deep reinforcement learning using grad-CAM: how AI plays atari games?. In *2019 IEEE Conference on Games (CoG)* (pp. 1-2). IEEE.
[29] Marmolejo-Saucedo, J.A. and Kose, U., 2022. Numerical grad-CAM based explainable convolutional neural network for brain tumor diagnosis. *Mobile Networks and Applications*, pp.1-10.
[30] Gorski, L., Ramakrishna, S. and Nowosielski, J.M., 2020. Towards grad-cam based explainability in a legal text processing pipeline. *arXiv preprint arXiv:2012.09603*.
[31] Apostolopoulos, I.D., Athanasoula, I., Tzani, M. and Groumpos, P.P., 2022. An Explainable Deep Learning Framework for Detecting and Localising Smoke and Fire Incidents: Evaluation of Grad-CAM++ and LIME. *Machine Learning and Knowledge Extraction*, *4*(4), pp.1124-1135.
[32] Lerma, M. and Lucas, M., 2022. Grad-CAM++ is Equivalent to Grad-CAM With Positive Gradients. *arXiv preprint arXiv:2205.10838*.
[33] Jung, Y.J., Han, S.H. and Choi, H.J., 2021. Explaining cnn and rnn using selective layer-wise relevance propagation. *IEEE Access*, *9*, pp.18670-18681.
[34] Ullah, I., Rios, A., Gala, V. and Mckeever, S., 2021. Explaining deep learning models for tabular data using layer-wise relevance propagation. *Applied Sciences*, *12*(1), p.136.
[35] Omeiza, D., Speakman, S., Cintas, C. and Weldermariam, K., 2019. Smooth grad-cam++: An enhanced inference level visualization technique for deep convolutional neural network models. *arXiv preprint arXiv:1908.01224*.
[36] Siddiqui, M.K., 2022. *QUANTIFYING TRUST IN DEEP LEARNING WITH OBJECTIVE EXPLAINABLE AI METHODS FOR ECG CLASSIFICATION* (Doctoral dissertation).
[37] Zafar, M.R. and Khan, N., 2021. Deterministic local interpretable model-agnostic explanations for stable explainability. *Machine Learning and Knowledge Extraction*, *3*(3), pp.525-541.
[38] Shi, S., Zhang, X. and Fan, W., 2020. A modified perturbed sampling method for local interpretable model-agnostic explanation. *arXiv preprint arXiv:2002.07434*.
[39] Schlegel, U., Vo, D.L., Keim, D.A. and Seebacher, D., 2022, February. Ts-mule: Local interpretable model-agnostic explanations for time series forecast models. In *Machine Learning and Principles and Practice of Knowledge Discovery in Databases: International Workshops of ECML PKDD 2021, Virtual Event, September 13-17, 2021, Proceedings, Part I*(pp. 5-14). Cham: Springer International Publishing.
[40] Zafar, M.R. and Khan, N.M., 2019. DLIME: A deterministic local interpretable model-agnostic explanations approach for computer-aided diagnosis systems. *arXiv preprint arXiv:1906.10263*.
[41] López, Y.A., Diez, H.R.G., Toledano-López, O.G., Hidalgo-Delgado, Y., Mannens, E. and Demeester, T., 2022, November. DLIME-Graphs: A DLIME Extension Based on Triple Embedding for Graphs. In *Knowledge Graphs and Semantic Web: 4th Iberoamerican Conference and third Indo-American Conference, KGSWC 2022, Madrid, Spain, November 21–23, 2022, Proceedings* (pp. 76-89). Cham: Springer International Publishing.
[42] Van den Broeck, G., Lykov, A., Schleich, M. and Suciu, D., 2022. On the tractability of SHAP explanations. *Journal of Artificial Intelligence Research*, *74*, pp.851-886.
[43] García, M.V. and Aznarte, J.L., 2020. Shapley additive explanations for NO2 forecasting. *Ecological Informatics*, *56*, p.101039.
[44] Chou, Y.L., Hsieh, C., Moreira, C., Ouyang, C., Jorge, J. and Pereira, J.M., 2022. Benchmark Evaluation of Counterfactual Algorithms for XAI: From a White Box to a Black Box. *arXiv preprint arXiv:2203.02399*.
[45] Joel, V., 2020. Explaining the output of a black box model and a white box model: an illustrative comparison.
[46] Brkan, M. and Bonnet, G., 2020. Legal and technical feasibility of the GDPR's quest for explanation of algorithmic decisions: of black boxes, white boxes and fata morganas. *European Journal of Risk Regulation*, *11*(1), pp.18-50.
[47] Calegari, R., Ciatto, G. and Omicini, A., 2020. On the integration of symbolic and sub-symbolic techniques for XAI: A survey. *Intelligenza Artificiale*, *14*(1), pp.7-32.
[48] Li, B., Qi, P., Liu, B., Di, S., Liu, J., Pei, J., Yi, J. and Zhou, B., 2023. Trustworthy ai: From principles to practices. *ACM Computing Surveys*, *55*(9), pp.1-46.
[49] Kaadoud, I.C., Fahed, L. and Lenca, P., 2021, August. Explainable AI: a narrative review at the crossroad of Knowledge Discovery, Knowledge Representation and Representation Learning. In *MRC 2021: Twelfth International Workshop Modelling and Reasoning in Context* (Vol. 2995, pp. 28-40). ceur-ws. org.
[50] Charte, D., Charte, F., del Jesus, M.J. and Herrera, F., 2020. An analysis on the use of autoencoders for representation learning: Fundamentals, learning task case studies, explainability and challenges. *Neurocomputing*, *404*, pp.93-107.
[51] Qian, K., Danilevsky, M., Katsis, Y., Kawas, B., Oduor, E., Popa, L. and Li, Y., 2021, April. XNLP: A living survey for XAI research in natural language processing. In *26th International Conference on Intelligent User Interfaces-Companion* (pp. 78-80).
[52] Jain, A., Ravula, M. and Ghosh, J., 2020. Biased models have biased explanations. *arXiv preprint arXiv:2012.10986*.
[53] Yeh, C.K., Hsieh, C.Y., Suggala, A., Inouye, D.I. and Ravikumar, P.K., 2019. On the (in) fidelity and sensitivity of explanations. *Advances in Neural Information Processing Systems*, *32*.
[54] Molnar, C., Casalicchio, G. and Bischl, B., 2021, February. Interpretable machine learning–a brief history, state-of-the-art and challenges. In *ECML PKDD 2020 Workshops: Workshops of the European Conference on Machine Learning and Knowledge Discovery in Databases (ECML PKDD 2020): SoGood 2020, PDFL 2020, MLCS 2020, NFMCP 2020, DINA 2020, EDML 2020, XKDD 2020 and INRA 2020, Ghent, Belgium, September 14–18, 2020, Proceedings* (pp. 417-431). Cham: Springer International Publishing.
[55] Wang, X., Ding, L., Ma, Z., Azizipanah-Abarghooee, R. and Terzija, V., 2020. Perturbation-based sensitivity analysis of slow coherency with variable power system inertia. *IEEE Transactions on Power Systems*, *36*(2), pp.1121-1129.
[56] Jhala, K., Pradhan, P. and Natarajan, B., 2020. Perturbation-based diagnosis of false data injection attack using distributed energy resources. *IEEE Transactions on Smart Grid*, *12*(2), pp.1589-1601.
[57] Rashki, M., Azhdary Moghaddam, M. and Miri, M., 2019. System-level reliability sensitivity analysis by using weighted average simulation method. *Quality and Reliability Engineering International*, *35*(6), pp.1826-1845.
[58] Xu, J. and Du, Q., 2020. Adversarial attacks on text classification models using layer-wise relevance propagation. *International Journal of Intelligent Systems*, *35*(9), pp.1397-1415.
[59] Kuppa, A. and Le-Khac, N.A., 2021. Adversarial xai methods in cybersecurity. *IEEE transactions on information forensics and security*, *16*, pp.4924-4938.
[60] Zhou, Z., Hooker, G., & Wang, F., 2021. S-lime: Stabilized-lime for model explanation. In Proceedings of the 27th ACM SIGKDD conference on knowledge discovery & data mining (pp. 2429-2438).
[61] Han, J., Brown, C., Chauhan, J., Grammenos, A., Hasthanasombat, A., Spathis, D., ... & Mascolo, C. , 2021. Exploring automatic COVID-19 diagnosis via voice and symptoms from crowdsourced data. In ICASSP 2021-2021 IEEE International Conference on Acoustics, Speech and Signal Processing (ICASSP) (pp. 8328-8332). IEEE.
[62] Dentamaro, V., Giglio, P., Impedovo, D., Moretti, L., & Pirlo, G., 2022. AUCO ResNet: An end-to-end network for Covid-19 pre-screening from cough and breath. Pattern Recognition, 127, 108656.
[63] J. Hlavnicka, R. Cmejla, J. Klempir, E. Ruzicka, and J. Rusz, 2019, "Acoustic tracking of pitch, modal, and subharmonic vibrations of vocal folds in Parkinson's disease and Parkinsonism," IEEE Access, vol. 7, pp. 150339–150354, doi: 10.1109/ACCESS.2019.2945874
[64] Srivastava, K., 2022. Deep learning techniques for music genre classification and feature importance. https://doi.org/10.36227/techrxiv.21265965

> REPLACE THIS LINE WITH YOUR MANUSCRIPT ID NUMBER (DOUBLE-CLICK HERE TO EDIT) <

19## APPENDIX A

**Algorithm 1** Multi Threaded EVIDENCE Theorem Implementation

```
// Define a thread for generating spectrograms
class SpectrogramThread
  function __init__(y, sr, filename='spectrogram.png')
    initialize y, sr, filename
    initialize result to None
  end function

  function run()
    try
      if y has 1 element then
        reshape y to 1D array
      end if
      S = generate Mel-spectrogram from y using sr
      S_DB = convert power spectrogram to decibel scale using S
      fig, ax = create figure and axes
      display S_DB using specshow from librosa
      remove axis from ax
      draw canvas of fig
      save fig as filename with tight bounding box, no padding, and transparent background
      close fig
      img = open filename as image
      if img mode is 'RGBA' then
        convert img to 'RGB'
      end if
      result = convert img to array
    catch exception as e
      print error message e
    end try
  end function

  function get_result()
    return result
  end function
end class

// Function for parallel processing
function PARALLEL(i, K, chunks, features, sound, chunk_len, samples)
  choices = select random elements from chunks with size = features
  zeros = array of zeros with the same shape as sound
  zero_chunks = split zeros into chunk_len parts
  zero_chunks = transpose zero_chunks
  zero_chunks = convert zero_chunks to floating-point numbers
  for j from 0 to length(choices) do
    ch = choices[j]
    zero_chunks[ch] = chunks[ch]
  end for
  sample = concatenate zero_chunks along axis 0
  sample = convert sample to floating-point numbers
  add sample to samples
end function

// Function for generating evidence
function EVIDENCE(sound, model, one_hot_instance_label, Y_train, chunks_hz=22, features=2,
    sampling_rate=16000, K=500, threshold=0.25)
  one_hot = reshape one_hot_instance_label to (1, length(one_hot_instance_label))
  chunk_len = chunks_hz
  chunks = split sound into chunk_len parts
  chunks = convert chunks to floating-point numbers
  chunks = transpose chunks
  instance_label = index of the maximum element in one_hot_instance_label
  samples = empty list
  scores = empty dictionary
  threads = empty list
  for i from 0 to K do
    create new thread with target = PARALLEL and args = (i, K, chunks, features, sound, chunk_len,
      samples)
    add new thread to threads
  end for
  for each thread in threads do
    start thread
  end for
  for each thread in threads do
    wait for thread to finish
  end for
  samples = convert samples to floating-point numbers
  probas = model.predict(samples)
  one_hots = empty list
  for i from 0 to length(probas) do
    add one_hot to one_hots
  end for
  one_hots = convert one_hots to array
  cce = calculate categorical cross entropy between one_hots and probas
  scaler = MinMaxScaler()
```

```
  cce = scaler.fit_transform(cce)
  for i from 0 to length(cce) do
    scores[cce[i]] = samples[i]
  end for
  reversed_keys = sorted keys of scores in ascending order
  sounds = empty list
  basethreshold = max(1, int(K * threshold))
  for f from 0 to min(basethreshold, length(reversed_keys)) do
    add scores[reversed_keys[f]] to sounds
  end for
  sounds = convert sounds to array
  if length(sounds) > 1 then
    mixed2 = mean of sounds along axis 0
    mixed2 = sum of mixed2 along axis 2
    arr = sum of mixed2 along axis 1
    pp = mean of arr
    indexes = indices where arr < pp
    mask = array of False with same length as mixed2
    set elements of mask at indexes to True
    for i from 0 to length of sound[1] do
      set sound[mask, i, :] to 0
    end for
  else
    sound = sounds[0]
  end if
  return sound
end function
```